\pdfoutput=1
\pdfoutput=1
\documentclass[11pt]{article}
\usepackage{cite}
\usepackage{EMNLP2022}
\usepackage{multirow}
\usepackage{times}
\usepackage{latexsym}
\usepackage{booktabs}
\usepackage{amsmath}
\usepackage{graphicx}
\usepackage[misc]{ifsym}

\usepackage[T1]{fontenc}

\usepackage[utf8]{inputenc}

\usepackage{microtype}

\usepackage{inconsolata}

\newenvironment{mybox}
    {
    \begin{tabular}{p{0.43\textwidth}}
    \\
    }
    { 
    \\
    \end{tabular} 
    }
\usepackage{xcolor}

\title{Do Charge Prediction Models Learn Legal Theory?}

\author{
    Zhenwei An$^{1,2} \thanks{\quad  Equal Contribution.}$, 
    Quzhe Huang$^{1,3*}$, 
    Cong Jiang$^{4,5}$, \\
    \Letter \textbf{Yansong Feng}$^{1,6}$\and
    \textbf{Dongyan Zhao}$^{1,5,6}$ \\
    $^1$Wangxuan Institute of Computer Technology, Peking University \\ $^2$School of Software \& Microelectronics, Peking University \\
    $^3$School of Intelligence Science and Technology, Peking University \\
    $^4$ Peking University Law School \ $^5$ Institute for Artificial Intelligence, Peking University \\
    $^6$ The MOE Key Laboratory of Computational Linguistics, Peking University \\
    {\tt \{anzhenwei,huangquzhe,jiangcong,fengyansong,zhaody\}} 
     {\tt @pku.edu.cn}
}

\begin{document}
\maketitle
\begin{abstract}
The charge prediction task aims to predict the charge for a case given its fact description. Recent models have already achieved impressive accuracy in this task, however, little is understood about the mechanisms they use to perform the judgment.
For practical applications, a charge prediction model should conform to the certain legal theory in civil law countries,  as under the framework of civil law, all cases  are judged according to certain local legal theories. In China, for example, nearly all criminal judges make decisions based on the Four Elements Theory (FET).
In this paper, we argue that trustworthy 
charge prediction models should take legal theories into consideration, and standing on prior studies in model interpretation, we propose three principles  for trustworthy models should follow in this task, which are \texttt {sensitive}, \texttt{selective}, and \texttt{presumption of innocence}.
We further design a new framework to evaluate whether existing charge prediction models learn legal theories. Our findings indicate that, while existing charge prediction models meet the \texttt{selective} principle on a benchmark dataset, most of them are still not \texttt{sensitive} enough and do not satisfy the \texttt{presumption of innocence}. Our code and dataset are released at \url{https://github.com/ZhenweiAn/EXP_LJP}.

\end{abstract}

\section{Introduction}

\begin{figure}[pt]
    \includegraphics[width=0.48\textwidth]{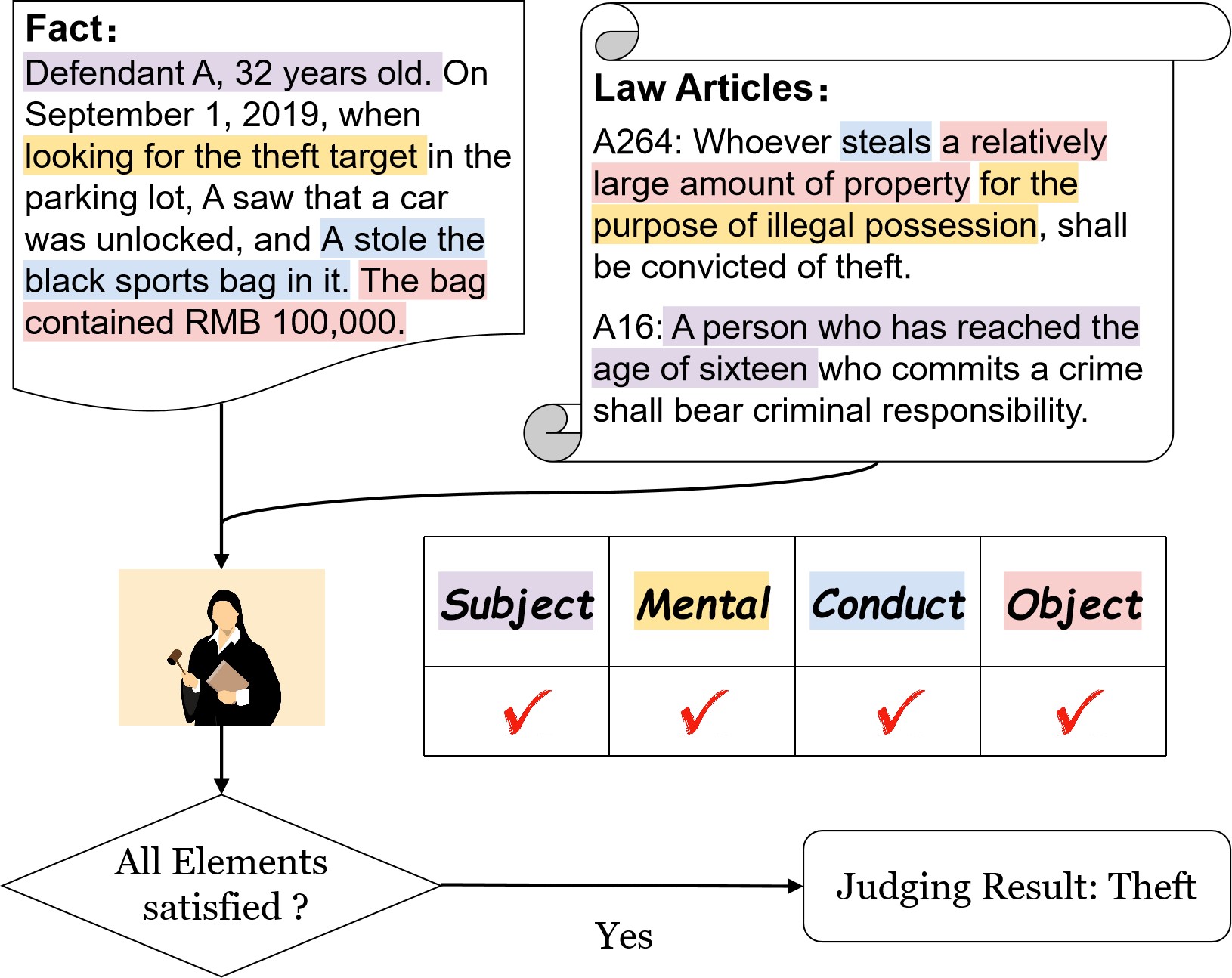}

    \caption{An example of accusing the defendant of \textit{Theft}. FET is the most dominant legal theory in China, which defines that a case must satisfy four criminal elements simultaneously to
constitute a crime}
    \label{fig:intro}
\end{figure}
The task of charge prediction is to determine appropriate charges, such as \textit{Fraud} or \textit{Theft}, for a case by analyzing its textual fact descriptions. Such a technique is beneficial for improving the efficiency of legal professionals, e.g., helping judges, lawyers, or prosecutors to distinguish similar charges and focus on discriminative features. But as an auxiliary tool in the legal domain, it should be used with great caution, in case of introducing undesirable unfairness~\cite{angwin2016machine}.

Most existing works formalize charge prediction as a text classification task~\cite{luo-etal-2017-learning,DBLP:conf/coling/HuLT0S18,zhong-etal-2018-legal}. Although recent advances in deep learning have demonstrated their excellent performance in predicting the charges~\cite{ijcai2019-567,RN126}, their reliability and interpretability are still under-explored. It is unknown whether the intrinsic decision mechanism of these models corresponds to the decision logic of human judges. 
Specifically, since most existing models are data-driven and all cases in the charge prediction dataset conform to local legal theories, it is necessary to figure out whether these charge prediction models learn their corresponding legal theories.

Previous studies have shown that trustworthy legal AI models are supposed to point out human-interpretable factors used in a decision \cite{atkinson2020explanation}. Besides, they should also explain how the changes in fact descriptions would change their decisions. Based on these discussions, 
we argue that a trustworthy charge prediction model should obey the following principles to conform to local legal theory and illustrate how they act in legal perspectives using FET, the most dominant legal theory in China \cite{wang2017criminal}, as an example:

1) \textbf{\texttt{Selective}}: be able to identify and concentrate on important parts of a case  when making decisions. In FET, the important parts are considered as \textit{criminal elements}.

2) \textbf{\texttt{Sensitive}}: be aware of the subtle distinctions between similar charges. When three of the four criminal elements in FET are identical for a pair of similar charges, a trustworthy model is expected to use the remaining criminal element to distinguish the similar charges. 

Apart from the prerequisites, which have been extensively explored in various domains, we can not ignore the presumption of innocence when focusing on a legal task. Presumption of innocence refers to the principle that any defendant is presumed innocent until proven guilty in a criminal trial, which is fundamental to protect human rights worldwide~\cite{tadros2004presumption}.
Taking this presumption into account, we propose an additional principle that a trustworthy charge prediction model should follow:
3) \textbf{\texttt{Presumption of innocence}}: always assume innocent unless sufficient requirements for a charge are met. In FET, \texttt{presumption of innocence} is guaranteed by checking all four criminal elements before making decisions.

In this paper, we propose a framework to evaluate whether a charge prediction model conforms to certain legal theory. Our framework consists of three components that evaluate the aforementioned principles respectively.
We first apply a probing task to measure whether models learn the skill of identifying criminal elements from fact descriptions, corresponding to the \texttt{selective} principle. The assumption here is  that if the model is capable of identifying criminal elements, the knowledge of such a skill should be reflected in its internal representations, which could be detected by a diagnostic model \cite{alt2020probing}.

The evaluation of the \texttt{sensitive} principle relies on a perturbation experiment, in which we modify the fact descriptions of confusing charges and check whether the model could detect the modifications. Specifically, for a pair of confusing charges, we rewrite the fact descriptions related to a certain criminal element and make the modified facts fulfill the requirements of the other charge.

If a model is \texttt{sensitive} enough, it should be capable of identifying these modifications and making different predictions for the original facts and the modified ones.
The final component evaluates whether models follow the \texttt{presumption of innocence} by checking the model's performance on incomplete fact descriptions.

Those incomplete facts are obtained by excluding all descriptions related to a specific criminal element from criminal descriptions. The models are expected to make innocent predictions for those incomplete fact descriptions, because they violate the requirements of FET that all the four criminal elements should be satisfied when judging guilty. 

We conduct experiments with  popular Chinese charge prediction models and the results indicate that, while existing charge prediction models meet the \texttt{selective} principle on our benchmark dataset, most of them are still not \texttt{sensitive} enough and do not satisfy the \texttt{presumption of innocence}.

Our contributions are four-folds: 

(1) We propose the first ever set of principles that a trustworthy charge prediction model should follow when conforming to certain legal theories.
(2) Based on these principles, we propose a new investigation framework to evaluate the trustworthiness of charge prediction models. (3) We supplement the current popular charge prediction dataset CAIL~\cite{Xiao2018CAIL2018AL} with innocent cases and provide sentence-level criminal elements annotation for a subset. (4) We examine existing Chinese charge prediction models using FET, the most widely used legal theory in China,  on the new benchmark, and find that most existing charge prediction models are not trustworthy enough, though they can achieve over 80\% prediction accuracy.

\section{The Charge Prediction Task}

\label{sec:crime_prediction}
 
Suppose the fact description of a case is a word sequence $ \mathbf{x} = \{x_1,x_2,\cdots,x_n\} $, where $n$ is the length of $\mathbf{x}$. Based on the fact description $\mathbf{x}$, the charge prediction task aims at predicting an appropriate charge $y \in Y$, where $Y$ is the potential charge set. 

To solve this task, previous works often use existing text classification models~\cite{8407150,RN53}, many of which are later improved by introducing legal knowledge~\cite{luo-etal-2017-learning,zhong-etal-2018-legal,ijcai2019-567}. More recently, pretrained language models have also been proven effective in this task~\cite{RN126}.

In our study, we select the following representative charge prediction models to evaluate whether they are trustworthy according to the specific legal theory, i.e., the FET in this case.

\paragraph{BiLSTM}  \citet{luo-etal-2017-learning} uses Bi-LSTM~\cite{DBLP:conf/naacl/YangYDHSH16} to encode fact descriptions and applies an attention mechanism to aggregate encoded word representations to obtain fact embedding, which is then used for classification.

\paragraph{TopJudge} TopJudge~\cite{zhong-etal-2018-legal} is a representative of those multitask learning models. During encoding, TopJudge employs CNN~\cite{DBLP:conf/emnlp/Kim14}  as the encoder to obtain fact embeddings. In decoding, it exploits a directed acyclic graph to capture the relationship among three sub-tasks, i.e., charge prediction, law article prediction, and term prediction, which are jointly optimized in a multitask framework.
\paragraph{FewShot} FewShot~\cite{DBLP:conf/coling/HuLT0S18} introduces discriminative attributes to distinguish confusing charges and provide additional knowledge for few-shot charges, which can stand for those models that introduce legal knowledge into the charge prediction task. It uses LSTM~\cite{10.1162/lstm} as the fact encoder and conducts charge prediction and attributes prediction afterward.
\paragraph{BERT} BERT~\cite{devlin-etal-2019-bert} is a strong baseline for many text classification tasks. We use the representation of [CLS] token for classification. 
\paragraph{Lawformer}\citet{RN126}  is a Longformer-based \cite{DBLP:journals/corr/abs-2004-05150} language model, which is pretrained on large-scale Chinese legal cases. We use it to encode the fact description and apply the classification based on the [CLS] token.

\subsection{The Four Elements Theory}
\label{ssec:FET}
Legal theories are the bases for judges to correctly determine charges, which define the method of analyzing cases. Judges are required to follow legal theories when making judgements~\cite{gao1993law}. If they do not, they might make decisions arbitrarily, which is a breach of human rights and freedom~\cite{wang2017criminal}.

In China, the Four Elements Theory~(FET) is the dominant legal theory for criminal trials. In practice, nearly all criminal judges use FET to justify their decisions~\cite{tang2011criminal}.
As a result, a trustworthy  Chinese charge prediction model should also conform to FET since they are trained based on the judgment documents which conform to the local legal theory, FET.

According to FET, a case must satisfy four criminal elements simultaneously to constitute a crime. The four criminal elements are: (1) the \emph{subject~(Sub)} refers to the person or organization who has committed the criminal offense and shall bear criminal crimes, (2) the \emph{object~(Obj)} refers to the person, thing, interest, or social relations protected by criminal law and jeopardised by criminal offence, (3) the \emph{conduct~(Con)} refers to harmful behaviors, and (4) the \emph{mental state~(Men)} is the mental state of the criminal subject when committing a crime, either \textit{intent} or \textit{negligence}.

For example, the four criminal elements of \textit{Theft} are as follows: (1) \emph{subject}: the general subject, that is, a person who has reached the age of criminal responsibility (16 years old in China), (2) \emph{object}: public or private property, (3) \emph{conduct}: the act of stealing a large amount of property or repeatedly stealing property, (4) \emph{mental state}: intent and with the purpose of illegal possession.

\section{Dataset}
\begin{table}[pt]
\small
\center
\begin{tabular}{c|cccc}
\toprule
   & Acc   & F1 & P & R \\ \midrule
TopJudge  & 82.7 & 60.6    & 67.5           & 59.2       \\ 
FewShot   & 82.9 & 71.7    & 75.9           &71.6
    \\
BiLSTM    & 82.4 & 59.8    & 65.7           & 58.9        \\ 
Bert      & 90.4 & 81.9    & 83.2           & 79.8        \\ 
Lawformer & 91.0 & 83.8    & 84.4           & 81.1        \\ 
\bottomrule
\end{tabular}
\caption{Charge Prediction results on CAIL-I, where Acc, F1, P, and R represent Accuracy, macro F1, macro precision, and macro recall, respectively.}
\label{tab:charge prediction results}
\end{table}

Existing charge prediction datasets, such as CAIL~\cite{Xiao2018CAIL2018AL}, have played a crucial role in the development of legal artificial intelligence research. However, they suffer from two limitations: (1) Lacking innocent cases. This violates the presumption of innocence, one of the most fundamental legal principles worldwide.  (2) Only containing coarse-grained annotations, such as charges and law articles, which cannot reveal how the judges analyze the cases.

To alleviate the two shortcomings, in this paper, we propose a new charge prediction dataset, CAIL-I, that adds innocent cases to the original CAIL. 
We further annotate whether a sentence is related to certain criminal elements in a subset of CAIL. We call this Sentence-level Criminal Elements dataset as SCE, which can be utilized to analyze whether a model conforms to FET.

\paragraph{Collecting Innocent Cases}
To obtain innocent cases, we first collect all non-prosecution cases from the Chinese Prosecutor's Website\footnote{www.12309.gov.cn}.
Among these non-prosecution cases, 

the real innocent cases take only small part. Many cases are not prosecuted for other reasons, such as the defendant died before prosecution. To identify the real innocent cases, we hire 2 law school graduate students to review the collected data case by case, and only kept the cases with truly innocent defendants. Finally, we obtain 462 innocent cases and add them into the CAIL training set, validation set, and test set at the ratio of 5:3:2 to form the new benchmark CAIL-I. We report the performance of existing charge prediction models on CAIL-I in Table~\ref{tab:charge prediction results}.

\paragraph{Annotating Criminal Elements}
\begin{table}[pt]
\small
\center
\begin{tabular}{c|ccccc|c}
\toprule
\multicolumn{1}{l|}{Charge} & Sub & Men  & Con  & Obj  & NA   & Cases \\ \midrule
TA                         & 102 & 228  & 258  & 163  & 833  & 100   \\
Rob                        & 109 & 185  & 435  & 153  & 673  & 98    \\
FS                         & 125 & 173  & 281  & 187  & 678  & 99    \\
Cor                        & 120 & 132  & 361  & 176  & 595  & 94    \\
MoF                        & 122 & 118  & 289  & 127  & 524  & 100   \\
MoPF                       & 122 & 133  & 318  & 135  & 571  & 97    \\
NH                         & 105 & 192  & 312  & 169  & 711  & 97    \\ \midrule
All                        & 805 & 1161 & 2254 & 1110 & 4585 & 685   \\ \bottomrule
\end{tabular}
\caption{Statistics of Sentence-level Criminal Elements dataset (SCE). Columns 2-5 show the number of  sentences involving different criminal elements, where NA means being related to none criminal elements. The last column indicates the number of cases corresponding to different charges. The abbreviations of criminal elements and charges are clarified in Section 3.}

\label{probing dataset}
\end{table}
Given a fact description and the corresponding charge label, annotators are asked to label each sentence with related criminal elements or NA when the sentence does not relate to any criminal elements. As a sentence might contain information about various criminal elements, it could be annotated with more than one label.
This annotation needs substantial legal knowledge involvement and the fine-grained scheme requires a huge workload, thus, it is impossible to annotate the whole CAIL-I datasets. 
To alleviate the burden of manual annotation, we choose 7 charges which are hard to distinguish in practice~\cite{Ouyang:99}. These confusing charges could help us better understand models' behavior under FET.
The 7 charges are \textit{Traffic accident (TA), Robbery (Rob), Forcible seizure (FS), Corruption (Cor), Misappropriation of funds (MoF), Misappropriation of public funds (MoPF)} and \textit{Negligent homicide (NH)}. We employed 2 paid graduate students from law schools as annotators and the inter-annotator Cohen’s Kappa is 0.64.
Table~\ref{probing dataset} shows the statistics of this Sentence-level Criminal Elements dataset (SCE).

\section{Selective Principle Checking}

Judges are required to extract criminal elements and filter less important information from fact descriptions, where the first step is to relate each sentence to its corresponding criminal elements. However, it is unclear whether existing charge prediction models are \texttt{selective} enough to relate sentences to criminal elements. 
To figure it out, we design a probing task to explore these models' ability to distinguish criminal elements of a charge.

\begin{figure}[pt]
\center
   \includegraphics[width=0.45\textwidth]{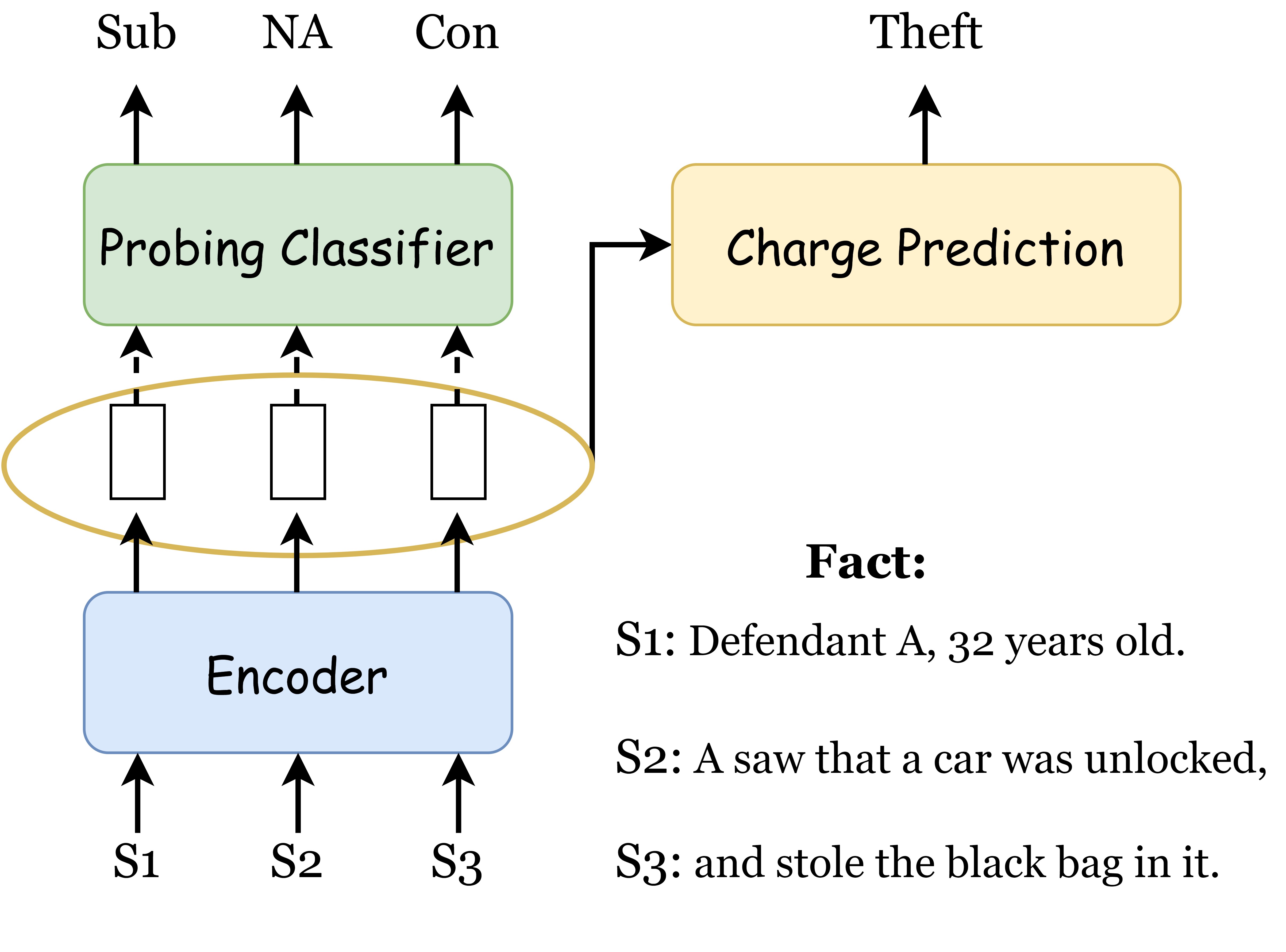}

    \caption{ Probing Setup. The dotted arrows emphasize that probing is applied to frozen encoders after training of charge prediction. Only the parameters of the linear classifier module are learned during probing. 
    }
    \label{fig:probing mechanism}
\end{figure}

Probing is a popular approach to model introspection, which trains a simple classifier -- a probe, to predict certain desired information from the latent representations learned by neural networks. High prediction performance is interpreted as evidence for the information being encoded in the representations and indicates that the information is what the neural networks rely on~\cite{saleh2020probing,alt2020probing}.
\begin{table*}
\centering
\resizebox{\linewidth}{!}{
\begin{tabular}{l|ccccccc}
\toprule
Models       & TA        & NH        & FS        & Rob       & MoF       & MoPF      & Cor       \\ \midrule
Random       & 14.3/25.3/19.3 & 18.1/28.9/22.3 & 16.3/23.6/19.5 & 22.3/33.4/26.9 & 18.7/28.5/22/6 & 20.2/30.8/24.4 & 21.1/32.1/25.5 \\
 \midrule
TopJudge     & 83.5/64.7/72.9 & 75.0/48.7/59.0 & 69.8/40.1/50.9 & 70.5/45.1/55.0 & 78.5/58.6/67.1 & 72.5/47.4/57.3 & 73.9/49.7/59.5 \\
FewShot     & 80.4/82.8/81.6 & 67.6/65.7/66.6 &64.3/57.6/60.7 & 59.5/52.9/56.0 & 72.2/65.4/68.6 & 64.1/52.5/57.7 & 63.1/53.4/57.8 \\
BiLSTM       & 86.6/82.8/84.7 & 72.3/65.6/68.8 & 68.8/54.4/60.8 & 69.6/53.3/60.4 & 80.8/62.3/70.4 & 72.1/51.2/59.9 & 71.9/55.8/62.8 \\
BERT         & 83.1/86.7/84.9 & 71.5/66.8/69.1 & 66.8/53.4/54.5 & 66.2/54.9/60.0 & 78.2/63.4/70.0 & 72.6/54.3/62.1 & 68.4/56.7/62.0 \\
Lawformer    & 84.5/83.2/83.8 & 72.8/66.6/69.5 & 66.7/55.5/60.6 & 67.2/55.7/60.9 & 80.8/66.6/73.0 & 70.9/52.1/60.0 & 73.1/60.1/66.0 \\ \midrule
ELMO*      & 85.6/84.4/84.9 & 75.2/67.1/70.9 & 70.9/55.5/62.2 & 67.5/53.7/59.9 & 81.0/71.0/75.7 & 72.1/57.7/64.1 & 73.2/62.2/67.3 \\
BERT*      & 85.8/85.8/85.8 & 74.4/66.3/70.1 & 73.2/56.3/63.6 & 71.9/57.4/63.8 & 80.9/69.2/74.6 & 79.3/57.2/66.5 & 78.1/58.7/67.0 \\
Lawformer* & 85.6/86.3/85.9 & 76.2/65.2/70.3 & 70.5/55.4/62.0 & 69.6/58.5/63.6 & 82.9/70.0/75.9 & 78.2/56.3/65.5 & 76.8/59.2/66.9 \\ \bottomrule
\end{tabular}
}
\caption{Precision/Recall/F1 of probing results for every charge in SCE. We report the average micro-metrics (\%) over 5 folds. The baseline performance is reported at the top and the performances of language models are shown at the bottom, indicated by *. 
}
\label{probing results}
\end{table*}

Our probing task is to examine whether the representations have encoded the type of criminal elements, which represents the model's ability to identify them, i.e., the \texttt{selectivity} of models. Figure~\ref{fig:probing mechanism} shows our probing setup. Specifically, we freeze encoders of charge prediction models and obtain the sentence representations of fact descriptions by the encoders. Then a linear classifier is trained on these representations to predict whether a sentence is related to certain criminal elements. This follows the same methodology used in previous works~\cite{saleh2020probing,alt2020probing}.

We apply mean pooling to get the sentence representation from the word embeddings encoded by the encoder of specific charge prediction models.
In order to explore how the charge prediction task influences models' ability to identify criminal elements, we conduct probing experiments on a few language models, %
including ELMO~\cite{peters-etal-2018-deep}, BERT, and Lawformer. The experiment is conducted on the SCE dataset.

\label{sec:criminal extraction extraction}
Table~\ref{probing results} shows the result of probing. We do 5-Fold Cross-Validation on SCE and report the average.
We use Random as a baseline, where we randomly assign the label to every sentence, based on the frequency of each label in the training set.  

\paragraph{Capacity of being \texttt{selective}}
As shown in Table~\ref{probing results}, all the charge prediction models outperform the baseline Random substantially. 
The good performance of charge prediction models in the probing task indicates that those models have learned the skill of identifying criminal elements and has the ability to distinguish them. In other words, existing charge prediction models are capable of being \texttt{selective}.

\paragraph{Effect of semantic information}

It is surprising to find that the language models, which are not finetuned on charge prediction, also perform well in the probing task. For example, BERT* achieves over 60\% micro-F1 scores for all the charges. This is because semantic information is enough for identifying criminal elements in many circumstances.
Taking the phrase ``\textit{car crush}'' as an example, it is easy to connect it with the \textit{conduct} element of \textit{Traffic accident}, when understanding this phrase describes a car hitting something. Even without legal knowledge, one will not consider ``car crush'' as introducing who was involved in an accident, i.e., the \textit{subject} element. Instead of the involvement of legal knowledge, understanding such phrases requires comprehending the semantic information of words or phrases, which has been successfully captured by language models like BERT.

\paragraph{Bias from shortcut} 

Another interesting finding is that models that have been trained for charge prediction perform worse in the probing task, e.g., BERT performs worse than BERT*. By comparing the predictions of these two groups of models, we discover that the performance drop in probing is due to the bias for particular patterns learnt by charge prediction models . 
During training, models learn that some patterns are highly correlated with specific charges, providing a shortcut for making judgments. 
Such a high correlation encourages the models to believe that those patterns are associated with specific criminal elements, a bias that ultimately results in the charge prediction models' worse performance in the probing task. 
This bias will lead to an incorrect association of sentences with certain criminal elements, which is evidenced by the fact that the decline of F1 is largely due to precision rather than recall, as shown in Table~\ref{probing results}.

\begin{mybox}
\small
A took advantage of the job convenience to keep 131,000 yuan of the company's business case, which \colorbox{yellow!50}{A had not paid back yet.}

\colorbox{blue!15}{[Conduct Element]}

\end{mybox}

\begin{figure*}[t]
\center
    \includegraphics[width=0.9\textwidth]{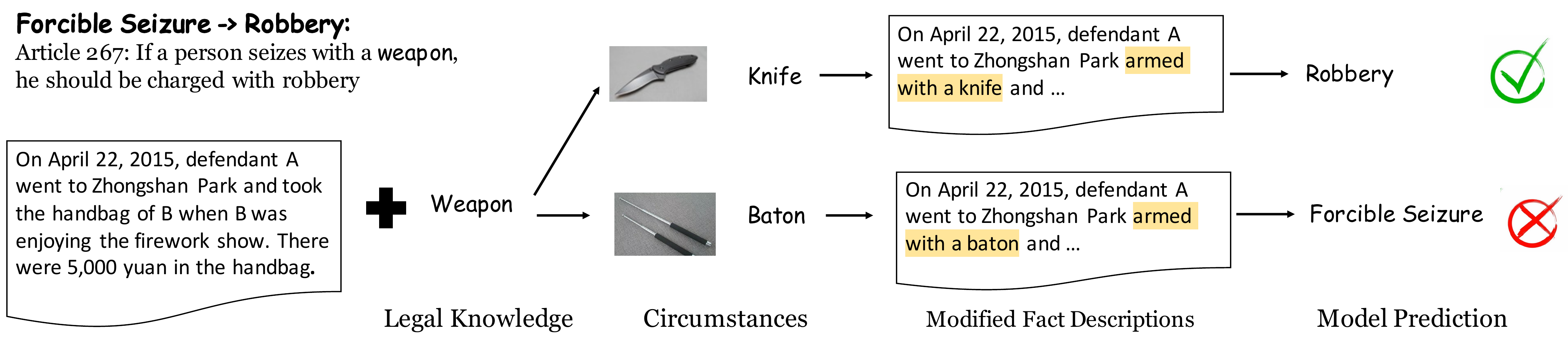}
    \caption{An example for \texttt{sensitive} principle checking.}
    \label{fig:perturbation_example}
\end{figure*}

The above case shows a detailed example of such bias. When learning charge prediction, Lawformer recognizes that the pattern of \textit{did not pay back money} is highly correlated with the \textit{conduct} element of \textit{Misappropriation of funds} without considering its context.

According to Chinese criminal law, the \textit{conduct} element of \textit{Misappropriation of funds} corresponds to the event where the defendants plunder money for more than three months and do not give the plundered money back before they are investigated. In most cases, the text pattern \textit{did not pay back money} implies the \textit{conduct} element of \textit{Misappropriation of funds}. However, in the case indicated by the following example, it does not imply any criminal element.

\begin{mybox}
\small
B misappropriated 140,000 yuan for personal costs. \colorbox{purple!20}{After being investigated},\colorbox{yellow!50}{B did not paid back the money} 
\colorbox{blue!15}{[Not Element]}
\end{mybox}

In this case, \textit{after being investigated} indicates that the crime had been completed and the legal authority began to investigate the crime. The act \textit{did not pay back money} that happened after investigation cannot be used for charge prediction,  hence does not belong to the \textit{conduct} element of \textit{Misappropriation of funds}.

\section{Sensitive Principle Checking}

A reliable charge prediction model should be \texttt{sensitive} to the subtle difference between the fact descriptions of confusing charges. In this section, we collect pairs of similar fact descriptions and evaluate whether existing models could recognize the difference.

Specifically, we select three groups of confusing charges where the charges in each group differ in only one criminal element from each other. According to the difference between the criminal elements of two charges, we modify the fact descriptions and make the modified ones meet the requirements of the other charge in the same group. We expect that a reliable model could recognize the distinctions and make different predictions for the original fact descriptions and the modified ones.

As shown in Figure~\ref{fig:perturbation_example}, the difference between \textit{Forcible Seizure} and \textit{Robbery} lies in their \textit{conduct} elements. ``armed with a weapon or not'' is the representative legal knowledge in distinguishing \textit{Robbery} from \textit{Forcible seizure}, as ``armed with a weapon'' could bring coercion to the victim, which is the \textit{conduct} element of \textit{Robbery}. Then, we add descriptions ``armed with a knife'' and ``armed with a baton'' to the fact descriptions of \textit{Forcible Seizure}. 
A trustworthy charge prediction model is expected to learn such legal knowledge during training and predict \textit{Robbery} for the modified fact descriptions.

For legal knowledge, like ``armed with weapons`` in Figure~\ref{fig:perturbation_example}, we design two specific circumstances, e.g., ``armed with a knife'' and ``armed with a baton'', where ``armed with a knife'' is much more common than the other. This design is to determine if the charge prediction models truly learn the legal knowledge, rather than simply remembering common textual patterns. If the model only recognizes the common circumstance and ignores the other, we believe the model does not learn that legal knowledge.

\begin{table}[ht]
\center
\resizebox{\linewidth}{!}{
\begin{tabular}{l|l|l}
\toprule
Changes                  & Legal Knowledge                   & Specific Circumstances                      \\ \midrule
\multirow{2}{*}{\textit{FS$\rightarrow$Rob}}  & armed with                      & armed with a baton                          \\ \cmidrule{3-3}
                         & weapon                          & armed with a knife                          \\ \midrule
\multirow{4}{*}{\textit{TFT$\rightarrow$Rob}} & \multirow{4}{*}{using violence} & spray the security        \\ 
						&								&   guards with pepper \\ \cmidrule{3-3}
                         &                                 & hurt pursuers with \\
                         &								&    a switchblade            \\ \midrule
\multirow{4}{*}{\textit{TA$\rightarrow$NH}}   &                    & on a road where the \\
						&				on non-public			& sewer is being repaired \\ \cmidrule{3-3}
                         & transport road                  & on a road closed for \\
                         &	&	construction           \\ \bottomrule
\end{tabular}
}
\caption{The legal knowledge and their corresponding circumstances used to modify the fact descriptions.}

\label{table:perturbation method}
\end{table}

Table~\ref{table:perturbation method} lists the three pairs of confusing charges and corresponding legal knowledge. 
For each pair, we randomly select 200 cases from the validation and test sets of CAIL-I and modify those fact descriptions with the two specific circumstances. 
The results are summarized in Table~\ref{perturbation results}.

\paragraph{Able to distinguish confusing charges?}
In most cases, charge prediction models still predict the original charge when the modified fact descriptions no longer satisfy the original one. Among three pairs of confusing charges, models perform best in distinguishing the \textit{conduct} element between \textit{Theft} and \textit{Robbery}, although the ratio of predicting the original charge still exceeds 50\%.  When it comes to 
\textit{Traffic accident} and \textit{Negligent homicide}, this ratio even reaches around 100\% for some models, indicating that these models totally fail to recognize the difference. 
It is surprising that FewShot does not perform well in this task, as it requires the model to pay explicit attention to several legal attributes. We think this is because their attributes are too coarse-grained and sparse. FewShot only designs 10 legal attributes for over 200 charges, and some legal knowledge, like ``on non-public transport road'', are not considered. 
The highly abstracted attributes may be useful in few-shot settings, but they cannot make the model more \texttt{sensitive}.
Overall, the poor performance in distinguishing confusing charges indicates that the existing  models are not \texttt{sensitive} enough.

\begin{table}[pt]
\small
\center
\setlength{\tabcolsep}{5pt}
\begin{tabular}{l|cc|cc|cc}
\toprule
Charge     & \multicolumn{2}{c|}{FS $\to$ Rob}  &\multicolumn{2}{c|}{TFT $\to$ Rob} & \multicolumn{2}{c}{TA $\to$ NH}    \\   
\midrule     
Circumstance      &      $\triangle$                  &  *               &    $\triangle$     &  * &    $\triangle$      &  * \\ 
\midrule
TopJudge  & 73.5                  & 45.5                 & 63.5 & 36.0    & 87.5 & 89.5   \\
FewShot  &93.5      &91.0    &87.5    &83.5 
&87.5 &84.0    \\
BiLSTM    & 82.5                  & 83.0                  & 77.5 & 47.0    & 92.0  & 89.5   \\
   BERT      & 88.0                   & 14.5                 & 88.0  & 26.0    & 96.0  & 79.5   \\
Lawformer & 84.0                   & 48.0                  & 59.5 & 33.5   & 98.0  & 97.5   \\ 
\bottomrule
\end{tabular}
\caption{The ratio of predicting the original charges after perturbations. The ``$\triangle$" refers to the more uncommon circumstance and ``*" refers to the common one.}
\label{perturbation results}
\end{table}

\paragraph{Textual patterns or legal knowledge?}
There are obvious discrepancies between models' capacities to recognize two distinct circumstances of the same legal knowledge. Taking BERT as an example, it cannot identify the uncommon circumstance ``armed with a baton'' for 88\% cases in the setting of \textit{Forcible seizure} and \textit{Robbery}. However, the models are very sensitive to the common one, ``armed with a knife'', with only 14.5\% cases maintaining the original prediction. The distinct performance suggests that charge prediction models are more likely to remember common textual patterns instead of understanding the legal knowledge necessary to discriminate between criminal elements of confusing charges.

\section{Presumption of innocence Checking}

\begin{table}[pt]
\small
\center
\begin{tabular}{lrrrr}
\toprule
{} &  Subject &  Mental &  Conduct &  Object \\
\midrule
BiLSTM    &    0.844 &   0.777 &    0.572 &   0.693 \\
TopJudge  &    0.772 &   0.666 &    0.574 &   0.644 \\
FewShot   &    0.826 &  0.753 &   0.619 &  0.740\\
BERT      &    0.920 &   0.866 &    0.704 &   0.826 \\
Lawformer &    0.924 &   0.880 &    0.712 &   0.841 \\
\bottomrule
\end{tabular}
\caption{The consistency of models' predictions between using the complete descriptions and the modified descriptions after removing the expressions related to one criminal element. }
\label{tab:internal_attack}
\end{table}

In China, FET guarantees the presumption of innocence by checking the completeness of all four criminal elements. The theory requires that only when all four criminal elements are satisfied, will a defendant be convicted of that charge. Based on this completeness checking, although A in the following example satisfied three criminal elements, A was innocent because A did not intend to occupy the phone and did not fulfill the requirement of the \textit{mental state} element.

\begin{mybox}
\small

A sat next to B on the bus. The wallet of B slipped out of B's pocket just before B got off the bus. A picked it up and \colorbox{yellow!50}{got off as far as possible to give it back to B}, but A did not find B. When B realize the wallet was lost, B called the police. Then the police arrested A on suspicious of the theft.

\end{mybox}

Ideally, a reliable model, which follows the \texttt{presumption of innocence}, should have the ability to check whether all the four criminal elements are satisfied and predict a case as innocent when its fact description lacks one or more criminal elements.
In this section, we explore whether existing charge prediction models have such an ability. 
Specifically, we generate cases lacking one criminal element by removing all the sentences related to that element in criminal fact descriptions and check whether the model could change its predictions when identifying such modifications. We use the SCE dataset to generate attack cases.

The ratio of predicting the same charge after removing one criminal element is shown in Table~\ref{tab:internal_attack}, and the confidence densities for predicting the original charge using the complete descriptions and the descriptions after removing the element-specific expressions are shown in Figure~\ref{fig:internal_attack}. The results of all charges and models are shown in Appendix. We can draw the following conclusions from Table~\ref{tab:internal_attack} and Figure~\ref{fig:internal_attack}:

\begin{figure}[pt]
\center
    \includegraphics[width=0.48\textwidth]{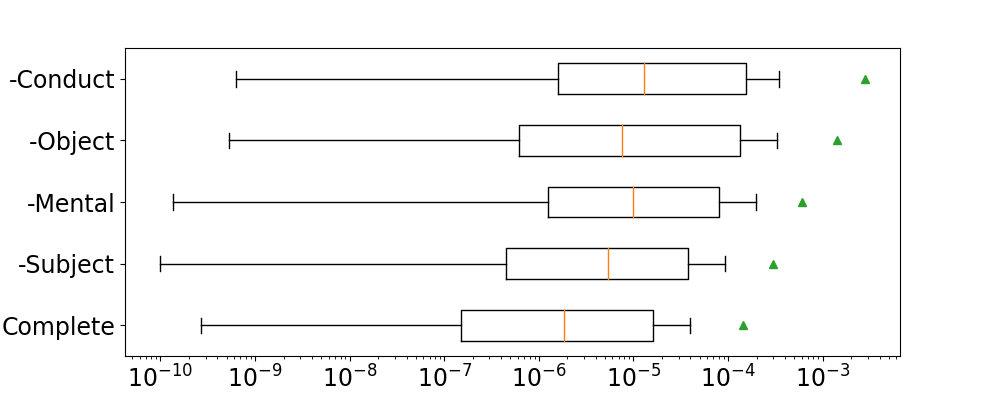}
    \caption{Boxplot describing Lawformer's prediction probability of innocent. "Complete" means complete fact descriptions. "-Subject" means fact descriptions with all sentences relevant to \textit{subject} element removed. So do "-Conduct", "-Object" and "-Mental".}
    \label{fig:innocent_box}
\end{figure}
\paragraph{Charge prediction models do not satisfy the \texttt{presumption of innocence}.} While charge prediction models are expected to recognize the absence of any criminal element, as shown in Table~\ref{tab:internal_attack}, all models stick to their predictions with incomplete fact descriptions most of the time.
Taking the \textit{subject} element as an example, when the descriptions related to it are deleted, TopJudge, the best-performing model, can maintain its predictions with a ratio of about 80\%, and for BERT and Lawformer, this ratio even exceeds 90\%. But it does not mean that these models completely ignore the absence of some elements. We discover that when a criminal element is removed, models improve the prediction probability of innocent, as shown in Figure~\ref{fig:innocent_box}. However, the improvement is insufficient to satisfy the \texttt{presumption of innocence}.

\paragraph{Which criminal elements gain more attention?}

As illustrated in Figure~\ref{fig:internal_attack}, Lawformer's confidence density changes significantly when the \textit{conduct} element is omitted. For the remaining three criminal elements, eliminating the relevant fact descriptions has little effect on the Lawformer's confidence in predicting the original charge, and in most cases, the confidence remains greater than 80\%, which is a very high level. The results of other models are shown in the Appendix and the results are similar as Lawformer's.

\section{Related Work}
\begin{figure}[pt]
\center
    \includegraphics[width=0.4\textwidth]{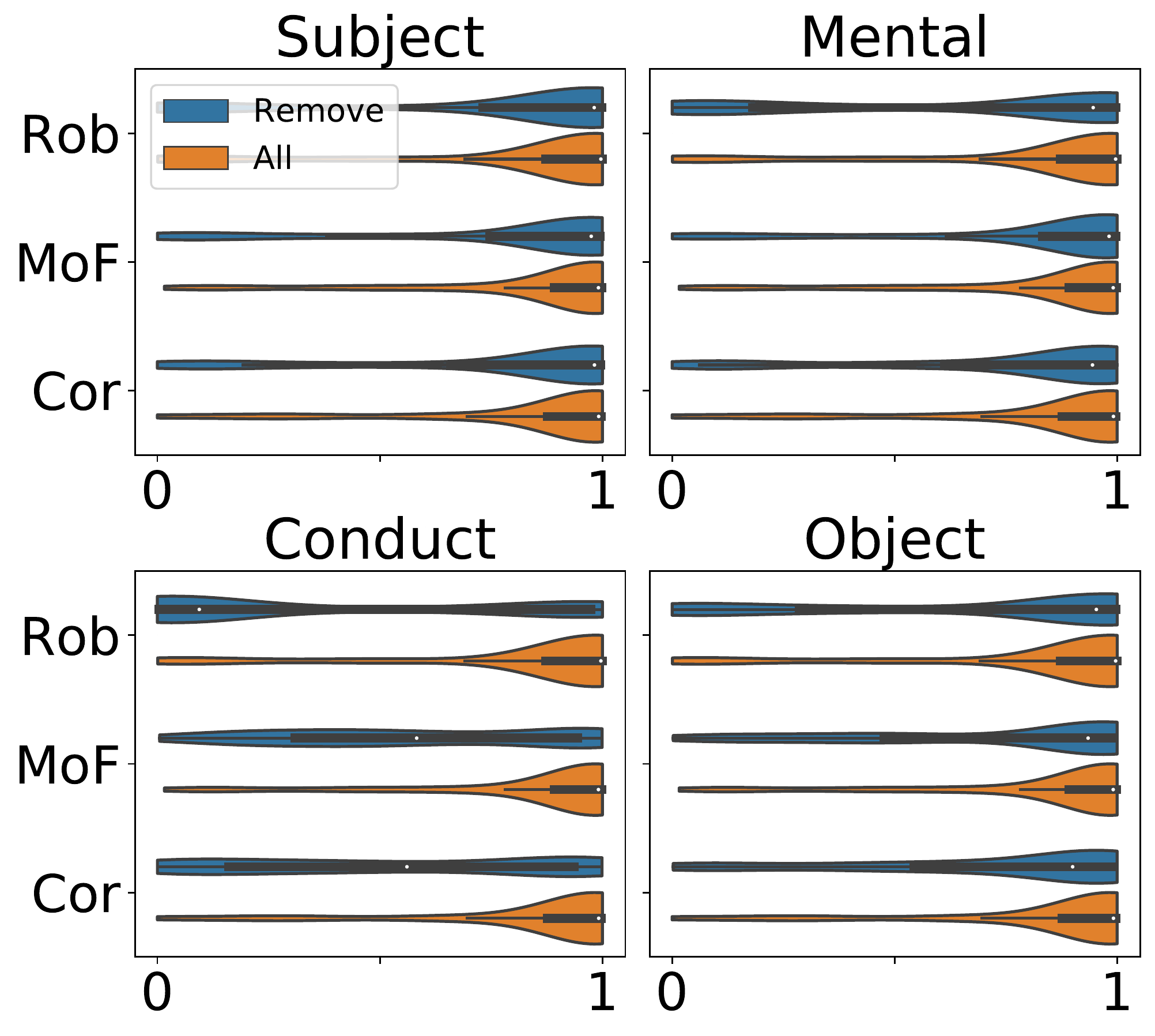}

    \caption{Confidence densities of predicting the original charges using the complete fact (orange) and using the fact after removing the descriptions related to a specific criminal element (blue). 
    }
    \label{fig:internal_attack}
\end{figure}
\paragraph{Probing} Probing is a popular method for model introspection, 
which associates the representations learned by the neural networks with properties of interests and examines the extent to which these properties can be recovered from the representations~\cite{ravichander2021probing,DBLP:conf/iclr/AdiKBLG17}. 

Previous works mainly focus on what linguistic properties  models have learnt~\cite{DBLP:conf/acl/BelinkovDDSG17,DBLP:conf/iclr/TenneyXCWPMKDBD19,DBLP:conf/emnlp/WarstadtCGPBNAB19,vulic2020probing}. There also has been research that employs probing to investigate the mechanisms that models used to perform certain tasks\cite{alt2020probing,saleh2020probing}. Although the probing method is widely used, as far as we know, no research has employed probing to explore whether neural networks could learn legal knowledge, like the Four Elements Theory in our paper. More discussion of probing could be found in \cite{belinkov2019analysis,belinkov2022probing}

Besides, we are not the first to analyze models by manipulating the input text. Many studies point out that minor changes in the text may bring unexpected results from the neural models, like Bert-Attack~\cite{li2020bert}. But little work has been done in the field of analyzing legal texts.

\paragraph{Interpretable Charge Prediction Models} When machine learning algorithms are put into practice for automated individual decision making, many legal experts demand \textit{right to explanation} \cite{right_to_explanation} for these algorithms, whereby users especially the losing party in a justification have the right to ask for an explanation of an algorithm decision that significantly affects them. Consequently, the reliability and interpretability of charge prediction models are of equal importance to their performance.

To improve the interpretability of charge predictions, several studies generate charge prediction alongside its supports. \citet{DBLP:conf/coling/JiangYLCM18} and \citet{liu19inproceedings} employ reinforce learning to derive rationales at the phrase level to explain the model's output. \citet{DBLP:conf/coling/HuLT0S18} and \citet{DBLP:conf/ictai/LiLYZF19} design certain attributes to help distinguish confusing charges. They train a classifier for each attribution and then make decisions depending on whether or not the fact descriptions satisfy those attributes. \citet{DBLP:conf/aaai/ZhongWTZ0S20} designs a series of questions and solves charge prediction by answering those questions, with each question corresponding to a certain attribute. \citet{liu2021everything} exhibits important
evidence for judgment with causal graphs and causal chains. \citet{Li21article} achieves multi-granularity inference of legal charges by obtaining the subjective and objective elements from the fact descriptions of legal cases. 

Although these strategies are more interpretable than those that just provide a charge, their reasoning procedure  may still violates FET. Some of them, such as \citet{DBLP:conf/ictai/LiLYZF19}, are solely concerned with the life-related \textit{object} element and disregard money-related \textit{object} element, thus cannot perform criminal element extraction for all charges. Some other works concentrate exclusively on a subset of the four criminal elements, omitting the others. \citet{DBLP:conf/coling/HuLT0S18}, \citet{DBLP:conf/aaai/ZhongWTZ0S20}, \citet{liu2021everything}, and\citet{Li21article} do not consider the \textit{subject} element, implying that they do not check whether all the elements are satisfied when making decisions, thus violate the \texttt{presumption of innocence}. As a result, the reliability of those models remains questionable.

\section{Conclusion and Future Work}
When applying artificial intelligence in the domain of law, not only do we expect a model to achieve high accuracy, but we also require the model to be trustworthy. Our work proposes three principles that a trustworthy model should follow in the charge prediction task based on both previous efforts on explanation and legal theories. According to the principles, we examine existing charge prediction models and our analysis shows that while they satisfy the \texttt{selective principle}, most models are not \texttt{sensitive} enough and do not satisfy the \texttt{presumption of innocence}.
We hope our discoveries will help the Artificial Intelligence and Law community better understand the mechanism of charge prediction models. We suggest the following directions for future work:

\begin{itemize}

    \item Extend current datasets with innocent cases to ensure models trained on them satisfy the \texttt{presumption of innocence}.
    \item Help models understand legal knowledge instead of identifying certain patterns.
    \item Design models which can perform completeness checking of all criminal elements before convicting the defendant of guilty.
\end{itemize}

\section*{Limitations}
Although this paper proposes principles of reliable charge prediction models based on legal theory and develops a framework for determining if a charge prediction model learns certain legal theory, we do not present a model which can actually adhere to these principles. It requires more exploration and research from the AI and Law community. In addition, the experiment designed to examine the \texttt{sensitive} principle requires substantial annotations  from legal experts, making it inconvenient in extending such method to other legal theories. 
Lastly, due to limited public criminal cases, we are only able to collect a subset of innocent instances.

\section*{Ethical Consideration}

\paragraph{Intended Use}
Our work could help the community of AI and Law better understand the mechanism of existing charge prediction models. We illustrate that the existing charge prediction models do not conform to the legal theory in China, and we call for using these models with more caution.

\paragraph{Misuse Potential}
Our work shows that existing charge prediction models are \texttt{selective} in our dataset, but that does not mean those models conform to the Four Element Theory. We think the existing charge prediction models could not replace judges and make predictions independently.

\section*{Acknowledgements}
This work is supported in part by NSFC (62161160339) and National Key R\&D Program of China (No. 2018YFC0831900). We would like to thank the anonymous reviewers for their helpful comments and suggestions; thank Chen Zhang, Xiao Liu and Weiye Chen for providing feedback on an early draft. For any correspondence, please contact Yansong Feng.

\bibliography{custom}
\bibliographystyle{acl_natbib}

\clearpage
\appendix

\section*{Appendix}
\label{sec:appendix}

\begin{figure}[h]
    \includegraphics[width=0.98\textwidth]{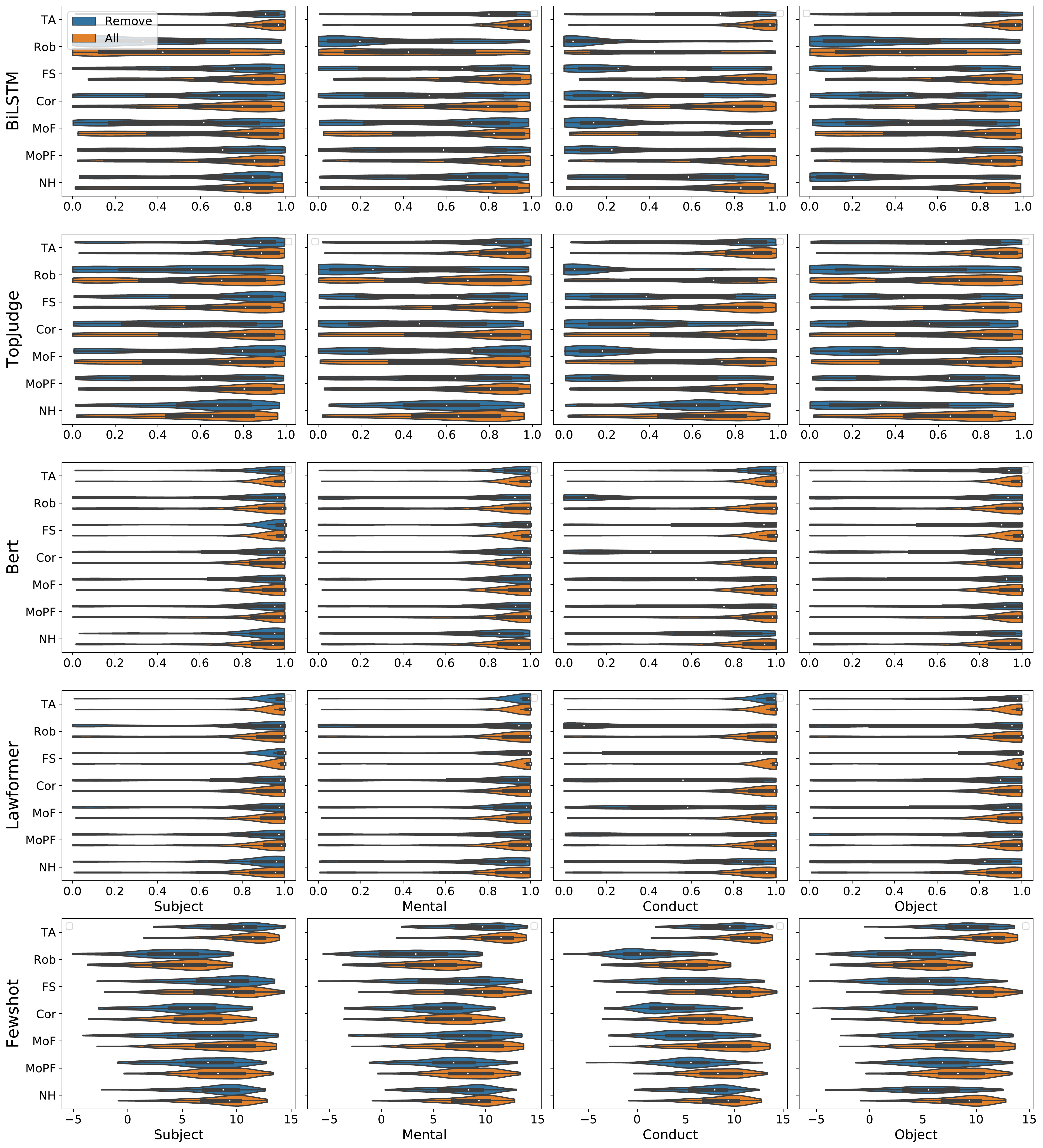}

    \caption{Confidence densities of predicting the original crimes using the complete fact (orange) and using the fact after removing the descriptions related to a specific criminal element (blue).}
    \label{fig:confidence}
\end{figure}

\end{document}